\documentclass[11pt]{article}
\usepackage[hyperref]{acl}
\usepackage{times}
\usepackage{latexsym}
\usepackage{graphicx}
\usepackage{booktabs}
\usepackage{amsmath}
\usepackage{multirow}
\usepackage{url}

\title{BatchDAG: LLM-Planned Execution Graphs for\\Scalable Ad-Hoc Analysis Over Enterprise Data}

\author{
  Anupreet Walia \\
  Brevian.ai \\
  \texttt{anu@brevian.ai}
}

\begin{document}
\maketitle

\begin{abstract}
Large language models (LLMs) excel at analyzing individual documents but break down on exhaustive, cross-entity analytical questions over enterprise-scale datasets due to context overflow, loss of per-entity attribution, and linear latency from sequential tool calls. We present BatchDAG, a system in which an LLM generates a typed directed acyclic graph (DAG) of operations---SQL queries, semantic searches, in-memory transforms, parallel fan-outs, and single-shot analyses---which a deterministic engine evaluates with topological-wave parallelism and structured JSON data flow. A key optimization, \emph{entity-aware batching}, groups rows by logical entity before fan-out, reducing LLM calls by up to 47$\times$. BatchDAG is not primarily an accuracy improvement over hand-optimized pipelines; rather, it is a general-purpose orchestration layer that replaces multiple hand-engineered workflows with a single system that generates the appropriate execution strategy from natural language. In controlled experiments on 12 transcript-heavy queries, BatchDAG (3.74/5) achieves quality comparable to an expert-designed pipeline (3.25/5) and significantly outperforms a ReAct agent (3.09/5, $p{<}0.01$), with superior provenance (77\% transcript evidence rate vs.\ 46--60\% for baselines). A controlled ablation shows structured JSON intermediates reduce hallucinations by 27\% versus prose summaries (paired $t$-test, $p{=}0.107$, $n{=}12$). The planner achieves 98.8\% valid-DAG rate across 300 planning calls. In production at Brevian.ai, BatchDAG processes queries over 50,000+ meetings in under 60 seconds, with measured per-query costs of \$0.02--\$0.24 at published GPT-5.1 pricing.
\end{abstract}

\section{Introduction}

Tool-augmented LLM agents have become the standard architecture for enterprise question-answering systems. In this paradigm, an LLM reasons about a user's question, selects tools (database queries, semantic search, API calls), processes the results, and generates an answer. Systems such as ReAct \citep{yao2023react}, Toolformer \citep{schick2023toolformer}, and commercial implementations in LangChain \citep{chase2022langchain} and LlamaIndex \citep{liu2022llamaindex} have demonstrated strong performance on queries that target individual entities or small document sets.

However, enterprise users increasingly ask a different class of question: exhaustive, cross-entity analytical queries that require processing hundreds to thousands of entities, each needing its own retrieval, contextual analysis, and per-entity attribution. Examples include:
\begin{itemize}
\item ``Did account executives start meetings with good discovery questions?'' (requires analyzing transcripts of every meeting)
\item ``Analyze every deal to see if security insurance was covered'' (requires per-deal transcript search and classification)
\item ``For each meeting, check if the key stakeholder negotiated on price'' (requires entity-level attribution across 3,000+ deals)
\end{itemize}

These queries expose three fundamental limitations of the single-agent loop:

\paragraph{Context window overflow.}
50,000 meetings $\times$ 25 top transcript results $=$ 1.25M tokens, exceeding even the largest commercial context windows.

\paragraph{Loss of per-entity attribution.}
Global top-$N$ search returns the $N$ most relevant results across \emph{all} entities. The system can report that a topic was discussed somewhere, but cannot attribute findings to specific deals.

\paragraph{Linear wall-clock time.}
Sequential tool calls mean latency grows proportionally with entity count, making large-corpus analysis impractically slow.

We present BatchDAG, a system that addresses these limitations by decomposing the problem into two phases: (1)~an LLM planner that generates a typed DAG of operations from a natural-language query, and (2)~a deterministic execution engine that evaluates the DAG with topological-wave parallelism, structured data flow, and entity-aware batching. The key insight is that most analytical queries can be decomposed into a small set of typed operations (SQL, search, transform, fan-out, analyze, compare), most of which require \emph{zero} LLM calls during execution. Only fan-out---the operation that applies LLM reasoning per entity batch---incurs LLM cost, and entity-aware batching minimizes that cost dramatically.

BatchDAG's primary contribution is not improving the quality of any single pipeline, but eliminating the need to design pipelines at all---replacing a collection of hand-engineered workflows with a single system that generates the appropriate execution strategy from natural language. In contrast to prior work that improves individual pipelines or agent reasoning, BatchDAG addresses a different problem: automatically selecting and composing the correct execution pipeline for each query from natural language.

Our contributions are:
\begin{enumerate}
\item A typed DAG formalism for decomposing ad-hoc analytical queries into composable operations with structured inter-step data flow.
\item An entity-aware batching algorithm that groups rows by logical entity before fan-out, achieving up to 47$\times$ reduction in LLM calls compared to row-level batching.
\item A goal-based planning prompt architecture that outperforms both exhaustive-rules and few-shot examples for generating correct DAGs.
\item A production deployment report covering cost, latency, and correctness over enterprise-scale data (50K+ meetings, 3K+ opportunities).
\item A controlled empirical evaluation demonstrating that automatically generated DAG pipelines achieve quality comparable to expert-designed baselines, with superior provenance (77\% transcript evidence rate) and 27\% fewer hallucinations through structured intermediates.
\end{enumerate}

\section{Related Work}

\subsection{Tool-Augmented LLM Agents}

ReAct \citep{yao2023react} interleaves reasoning and action steps in a single LLM loop. Toolformer \citep{schick2023toolformer} fine-tunes language models to insert API calls. Both operate on single queries and do not address the entity-level parallelism required for cross-corpus analysis. Commercial frameworks including LangChain, LlamaIndex, and Semantic Kernel provide tool-calling abstractions but delegate execution planning to the LLM at each step, inheriting the sequential bottleneck.

\subsection{Query Planning and Decomposition}

Least-to-Most prompting \citep{zhou2023leasttomost} decomposes complex questions into sub-questions but generates natural-language intermediate results, losing structured composability. Plan-and-Solve \citep{wang2023planandsolve} generates multi-step plans but executes them sequentially within a single context. Self-Discover \citep{zhou2024selfdiscover} selects reasoning modules for task composition but targets single-instance reasoning rather than data-parallel analysis. SQL generation approaches such as DIN-SQL \citep{pourreza2023dinsql} and others \citep{gao2023texttosql} handle structured queries but cannot combine SQL with unstructured retrieval and LLM-based analysis in the same execution graph.

\subsection{Map-Reduce for LLM Workloads}

Map-Reduce patterns have been applied to LLM summarization and document analysis. However, these approaches use fixed two-stage pipelines (map all documents, then reduce) and do not support the heterogeneous operation graphs (SQL $\to$ search $\to$ transform $\to$ fan-out $\to$ analyze) required for complex analytical queries. BatchDAG generalizes the Map-Reduce pattern by allowing arbitrary DAG structures with typed operations and structured data flow between steps.

\section{System Design}
\label{sec:design}

\begin{figure}[t]
\centering
\includegraphics[width=\columnwidth]{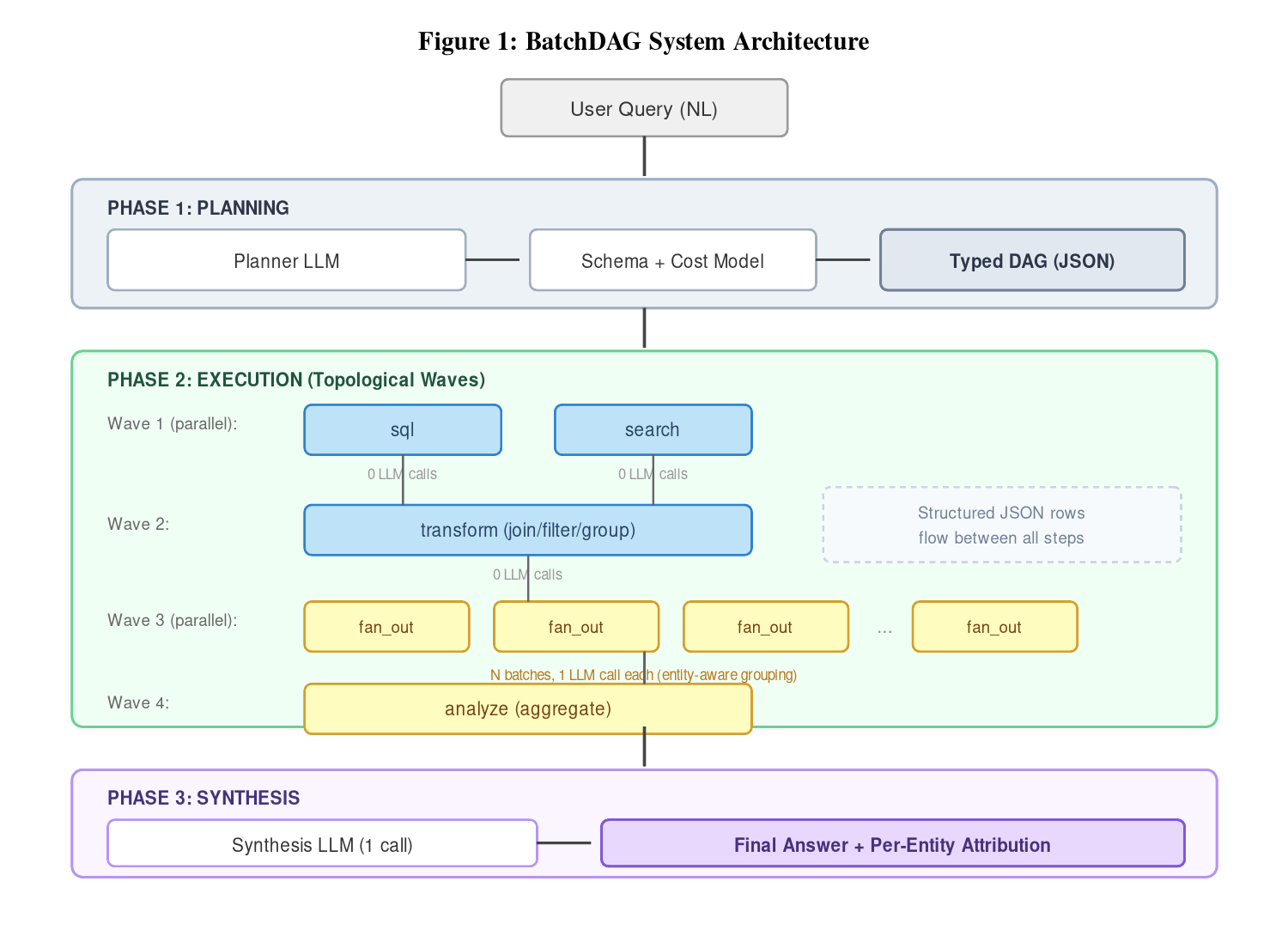}
\caption{BatchDAG system architecture. Phase~1: an LLM planner generates a typed DAG. Phase~2: a deterministic engine executes steps in topological waves. Phase~3: a synthesis LLM compiles results with citations.}
\label{fig:architecture}
\end{figure}

BatchDAG operates in three phases: planning, execution, and synthesis. Figure~\ref{fig:architecture} illustrates the overall architecture.

\subsection{Architecture Overview}

A user query enters the system and initiates a BatchDAG workflow on a durable execution engine. The workflow proceeds through three phases:

\textbf{Phase 1 (Planning):} A planner LLM receives the user query along with a schema description of available operations and data sources. It generates a typed DAG of steps, each with explicit input specifications declaring which prior step's output to consume and how to join or filter it.

\textbf{Phase 2 (Execution):} The execution engine performs a topological sort of the DAG and executes steps in waves. Steps within the same wave (no mutual dependencies) execute in parallel. SQL, search, transform, and compare steps are deterministic and require zero LLM calls. Fan-out steps spawn parallel child tasks that process entity batches concurrently. Analyze steps make a single LLM call on aggregated data.

\textbf{Phase 3 (Synthesis):} A synthesis LLM compiles all step results into a coherent final answer with per-entity citations and attribution.

\subsection{Typed Operation DAG}

Each step in the DAG has a type drawn from a fixed set of six operations. Table~\ref{tab:steptypes} summarizes the step types, their computational backends, and their LLM cost.

\begin{table}[t]
\centering
\small
\begin{tabular}{@{}llcl@{}}
\toprule
\textbf{Type} & \textbf{Description} & \textbf{LLM} & \textbf{Backend} \\
\midrule
\texttt{sql}       & Parameterized DB query  & 0         & Relational DB \\
\texttt{search}    & Semantic/keyword search  & 0         & Vector + reranker \\
\texttt{transform} & Filter, join, group, sort & 0        & Pure Python \\
\texttt{fan\_out}  & Per-entity LLM analysis   & $N$       & LLM API \\
\texttt{analyze}   & Single aggregation call   & 1         & LLM API \\
\texttt{compare}   & Differential analysis     & 0         & Pure Python \\
\bottomrule
\end{tabular}
\caption{BatchDAG step types. Four of six types require zero LLM calls during execution.}
\label{tab:steptypes}
\end{table}

The critical design decision is that four of six step types (\texttt{sql}, \texttt{search}, \texttt{transform}, \texttt{compare}) require zero LLM calls during execution. Only \texttt{fan\_out} and \texttt{analyze} invoke the LLM, and \texttt{analyze} is always a single call. This means that for queries answerable by SQL + transform + analyze, the total LLM cost after planning is exactly one call.

\subsection{Structured Inter-Step Data Flow}

Steps pass structured JSON rows between them, never prose summaries. This is the single most important architectural decision in BatchDAG. Each step declares its input via an \texttt{InputSpec}:

\begin{itemize}
\item \textbf{Direct reference:} \texttt{\{step: 1, key: "meeting\_id"\}} extracts a specific column from a prior step's output.
\item \textbf{Merge join:} \texttt{\{merge: [\{step: 1, key: "id"\}, \{step: 3, key: "meeting\_id"\}]\}} performs a left-join between two prior steps.
\item \textbf{Dependency resolution:} \texttt{depends\_on: [1, 2]} auto-resolves from the first dependency with available data.
\end{itemize}

When we initially allowed the LLM to summarize intermediate results in natural language, downstream steps hallucinated data and lost attribution. Structured rows are less expressive but fully composable: they support real database-style joins, filters, and grouping between steps, and they preserve the provenance chain from source data to final answer.

\subsection{Entity-Aware Batching}

The \texttt{fan\_out} step is the only operation where cost scales with entity count. Naive row-level batching produces catastrophic inefficiency. Consider a query that processes 121 meetings, each with approximately 48 transcript rows, yielding 5,824 total rows. With a batch size of 5 rows:

\textbf{Row-level batching:} $\lceil5824/5\rceil = 1{,}165$ batches. The same meeting is analyzed 5--50 times, each time with an incomplete fragment of its transcript.

\textbf{Entity-aware batching:} Group by \texttt{meeting\_id} first $\to$ 121 entities. $\lceil121/5\rceil = 25$ batches. Each batch receives 5 complete meetings with all their transcript rows. This achieves a \textbf{47$\times$ reduction} in LLM calls while ensuring each entity receives holistic analysis with full context.

\begin{figure}[t]
\centering
\includegraphics[width=\columnwidth]{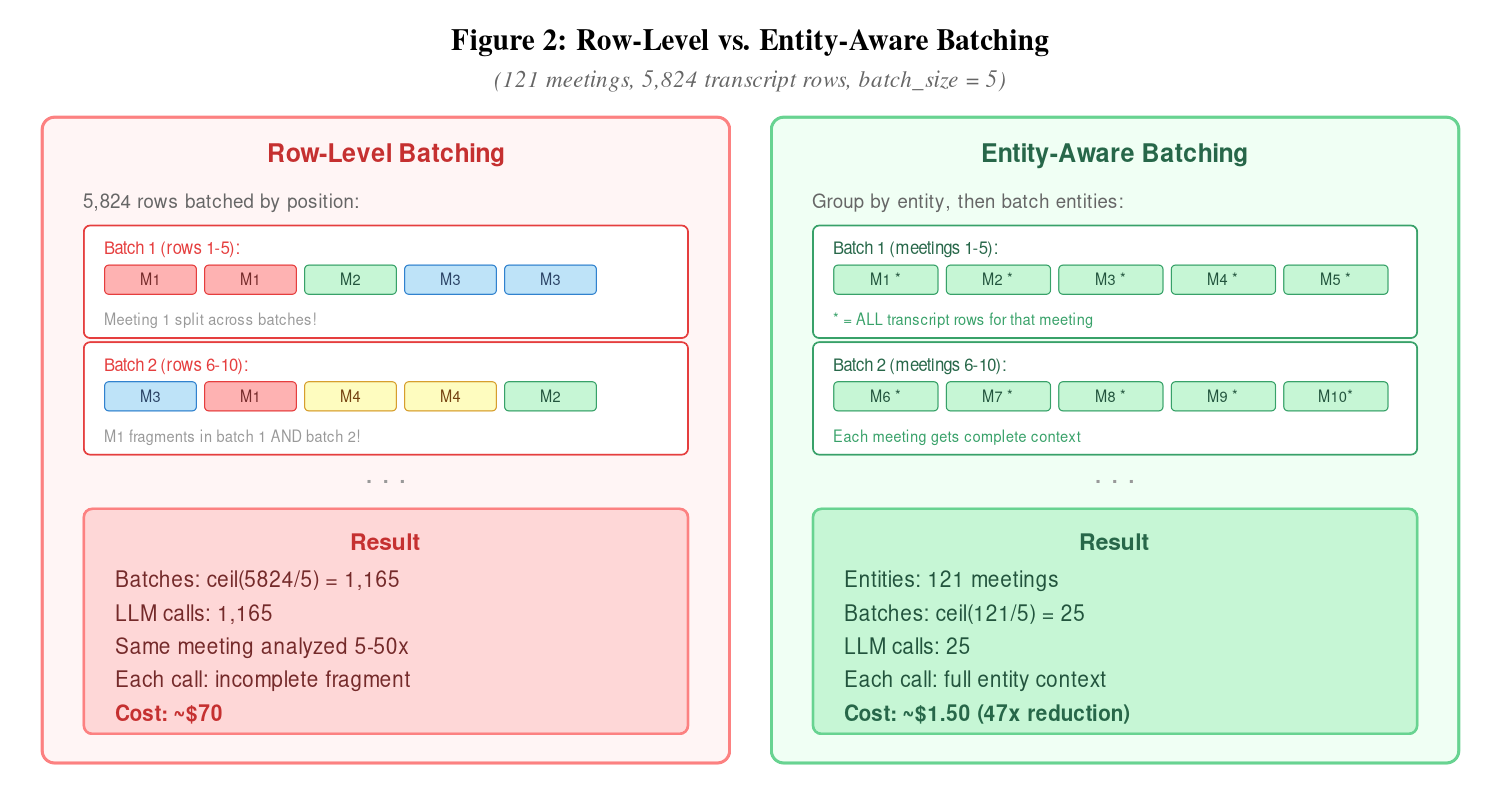}
\caption{Row-level vs.\ entity-aware batching. Entity-aware batching groups by the logical unit of analysis (meeting), achieving 47$\times$ fewer LLM calls with complete per-entity context.}
\label{fig:batching}
\end{figure}

Figure~\ref{fig:batching} illustrates the difference.

\subsection{Topological-Wave Execution}

The execution engine computes the topological ordering of the DAG and groups steps into waves of mutually independent operations. Steps within a wave execute in parallel; the engine blocks between waves. For a typical 5-step analytical query:

\begin{itemize}
\item \textbf{Wave 1:} [\texttt{sql}, \texttt{search}] $\to$ parallel data retrieval
\item \textbf{Wave 2:} [\texttt{transform}] $\to$ join results from wave 1
\item \textbf{Wave 3:} [\texttt{fan\_out}] $\to$ parallel entity-batch LLM calls
\item \textbf{Wave 4:} [\texttt{analyze}] $\to$ synthesize findings
\end{itemize}

Fan-out steps spawn child tasks on the durable execution engine, enabling per-batch retry semantics. If a single batch fails (LLM timeout, rate limit), only that batch is retried---not the entire fan-out.

\subsection{Storage Tiers}

Inter-step data is stored according to volume. Table~\ref{tab:storage} summarizes the three tiers.

\begin{table}[t]
\centering
\small
\begin{tabular}{@{}llll@{}}
\toprule
\textbf{Tier} & \textbf{Threshold} & \textbf{Storage} & \textbf{TTL} \\
\midrule
Small  & ${<}$10K rows & Inline params & Ephemeral \\
Medium & 10K--100K rows & In-memory KV & 2 hours \\
Large  & ${>}$100K rows & DB temp table & Session \\
\bottomrule
\end{tabular}
\caption{Storage tier selection for inter-step data.}
\label{tab:storage}
\end{table}

\section{LLM-Based DAG Planning}
\label{sec:planning}

\subsection{Goal-Based Prompting}

The planner prompt went through three iterations during development, yielding an important empirical finding about prompt architecture for code-generation tasks:

\textbf{Iteration 1: Exhaustive rules.} Listed every possible plan pattern explicitly. The LLM still generated unexpected structures not covered by the rules.

\textbf{Iteration 2: Few-shot examples.} Provided five example query$\to$DAG pairs. The LLM copied examples even when they did not fit the query, producing over-engineered 12-step plans with unnecessary SQL derivation.

\textbf{Iteration 3: Goal-based (deployed).} Describes what each operation type does, its cost characteristics, and its data model. The LLM reasons from first principles about which operations to compose. This approach is more robust to novel queries because the LLM reasons about the problem rather than pattern-matching against examples.

The planner receives a schema description of available data sources, the set of available operation types with their parameter specifications, and explicit cost annotations (``\texttt{fan\_out} is THE expensive one''). It outputs a JSON DAG with typed steps, input specifications, and dependency declarations.

\subsection{Defending Against LLM Over-Specification}

The planner LLM frequently generates specification fields that do not exist in the system's type definitions (e.g., \texttt{output\_mode}, \texttt{max\_tokens}). Rather than iteratively constraining the prompt, we implemented a safe-specification filter that strips unknown fields at construction time. The LLM can hallucinate arbitrary parameters; the system ignores what it does not understand. This makes the system robust to model upgrades and prompt variations without requiring prompt-level iteration.

\subsection{Output Format Constraints}

A critical practical finding: LLM output format matters more than prompt quality. The planner initially returned markdown-wrapped JSON, causing parse failures and silent fallback to a single-step plan. Forcing structured JSON output mode fixed this immediately. However, structured output mode wraps array responses in object containers, requiring an unwrap step to extract the actual results. We recommend always testing the raw bytes an LLM returns, not what the prompt implies it should return.

\section{Production Deployment}
\label{sec:production}

\subsection{Deployment Context}

BatchDAG is deployed in production at Brevian.ai, an enterprise AI platform for sales intelligence. The system serves analytical queries over organizational data including meeting transcripts, CRM records, structured sales methodology extractions, stakeholder data, and knowledge base documents. A typical enterprise organization in the system contains approximately 50,000 meetings, 3,000 opportunities, and 50,000 stakeholder records, with an average of 17 meetings per deal and 48 transcript rows per meeting.

\subsection{Cost and Latency}

Table~\ref{tab:prodcost} reports cost and latency across query categories. Costs are computed from actual GPT-5.1 token counts captured during experiments at published API pricing (\$1.25/1M input, \$10/1M output).\footnote{All costs are computed from actual GPT-5.1 token counts captured in experiment run metadata, multiplied by published API pricing at the time of writing (\$1.25/1M input tokens, \$10/1M output tokens). The ``Full corpus'' row is extrapolated to 3K-opportunity scale from measured per-entity costs.}

\begin{table}[t]
\centering
\small
\begin{tabular}{@{}lcccc@{}}
\toprule
\textbf{Query Category} & \textbf{Steps} & \textbf{LLM} & \textbf{Cost} & \textbf{Time} \\
\midrule
SQL-only            & 2 & 2   & \$0.02      & ${\sim}$5s \\
Search + analyze    & 3 & 3   & \$0.03      & ${\sim}$10s \\
Fan-out (100 mtgs)  & 5 & 22  & \$0.24      & ${\sim}$30s \\
Full corpus (3K deals) & 5 & 62 & ${\sim}$\$4 (ext.) & ${\sim}$60s \\
\bottomrule
\end{tabular}
\caption{Production cost and latency by query category. The ``Full corpus'' row is extrapolated from measured per-entity costs.}
\label{tab:prodcost}
\end{table}

The cost structure demonstrates the effectiveness of the typed operation approach: SQL-only queries cost \$0.02 (two LLM calls: plan + synthesis), while full-corpus analysis is extrapolated to stay under \$4 because only \texttt{fan\_out} batches invoke the LLM during execution. The 60-second wall time for 3,000-deal analysis compares favorably to the theoretical minimum of a sequential agent approach: 3,000 entities $\times$ ${\sim}$2s per tool call $=$ ${\sim}$100 minutes.

\subsection{Entity-Aware Batching Impact}

Table~\ref{tab:batching} quantifies the impact of entity-aware batching on a representative query over 121 meetings with 5,824 transcript rows.

\begin{table}[t]
\centering
\small
\begin{tabular}{@{}lcc@{}}
\toprule
\textbf{Metric} & \textbf{Row-Level} & \textbf{Entity-Aware} \\
\midrule
Total rows                & 5,824    & 5,824 \\
Batches (size=5)          & 1,165    & 25 \\
LLM calls                 & 1,165    & 25 \\
Reduction factor           & 1$\times$ & 47$\times$ \\
Per-entity context          & Fragmented & Complete \\
Estimated cost             & ${\sim}$\$70 & ${\sim}$\$1.50 \\
\bottomrule
\end{tabular}
\caption{Row-level vs.\ entity-aware batching on a representative query (121 meetings, 5,824 transcript rows).}
\label{tab:batching}
\end{table}

Beyond the 47$\times$ cost reduction, entity-aware batching qualitatively changes the analysis: each entity receives its complete context, enabling holistic classification rather than fragment-level guessing.

\section{Empirical Evaluation}
\label{sec:evaluation}

We evaluate BatchDAG across four dimensions: planner correctness, end-to-end answer correctness, comparative answer quality, and hallucination control. All experiments use a development organization containing 46 meetings, 62 open deals, and 32 accounts in the Brevian platform.

\subsection{Planner Correctness}

We curated a benchmark of 100 diverse natural-language queries spanning five categories: SQL-only lookups, search-only retrieval, simple fan-out, complex fan-out, and edge cases. Each query was run through the planner 3 times (300 total calls).

\textbf{Results.} Of 300 planner calls, 258 succeeded (42 failed due to API errors, not planner logic). Of the 258 successful plans, 255 produced valid, executable DAGs---a validity rate of \textbf{98.8\%}. The three invalid plans contained schema errors (referencing non-existent columns). Across the three runs, per-run validity rates were 98.8\%, 98.8\%, and 98.9\%, demonstrating high consistency. The planner achieved 100\% validity on SQL-only and search-only categories; all three failures occurred on complex fan-out queries requiring multi-source joins.

\subsection{End-to-End Answer Correctness}

We created 13 queries with gold-standard answers verifiable against the development database. Query types included exact counts, list retrieval, and aggregations.

\textbf{Results.} BatchDAG produced correct answers for 12 of 13 queries (\textbf{92.3\%} accuracy). The single incorrect answer was a count query where BatchDAG returned 128 vs.\ the gold answer of 108, likely due to a filter boundary condition in the generated SQL.

\subsection{Comparative Evaluation}

We designed 12 transcript-heavy queries (TX01--TX12) requiring both SQL metadata and specific transcript evidence. We compare three system architectures, each run 3 times (36 query-runs per system):

\begin{itemize}
\item \textbf{S5b: Enhanced Fan-out.} SQL $\to$ per-entity search + GPT-5.1 analysis $\to$ synthesis. A hardcoded three-phase pipeline representing the strongest baseline.
\item \textbf{S3: Intelligence (ReAct Agent).} Brevian's existing ReAct-style agent with access to SQL, search, and analysis tools.
\item \textbf{S8: BatchDAG.} The system described in this paper.
\end{itemize}

Answer quality was scored by a GPT-5.1 judge (temperature=0.0, JSON mode) on six dimensions plus a holistic overall score (1--5).

\begin{table}[t]
\centering
\small
\begin{tabular}{@{}lccccc@{}}
\toprule
\textbf{System} & \textbf{Overall} & \textbf{SD} & \textbf{TxEvid.} & \textbf{Latency} \\
\midrule
S5b: Enh.\ FO   & 3.25 & 0.90 & 46\%  & ${\sim}$207s \\
S3: Intelligence & 3.09 & 1.31 & 60\%  & ${\sim}$102s \\
S8: BatchDAG     & 3.74 & 1.15 & 77\%  & ${\sim}$94s \\
\bottomrule
\end{tabular}
\caption{Three-system comparison on 12 transcript-heavy queries ($\times$3 runs). S5b scores exclude Run~3 (infrastructure failure).}
\label{tab:comparison}
\end{table}

\subsection{Statistical Analysis}

Table~\ref{tab:stats} reports paired $t$-tests and bootstrap 95\% confidence intervals.

\begin{table}[t]
\centering
\small
\begin{tabular}{@{}lccccc@{}}
\toprule
\textbf{Comparison} & $\Delta$ & $t$ & \textbf{95\% CI} & $d$ & $n$ \\
\midrule
BD vs.\ Intel.       & +0.68 & 2.76\rlap{**} & [+0.18, +1.17] & 0.48 & 34 \\
BD vs.\ Enh.\ FO     & +0.48 & 1.56          & [$-$0.16, +1.12] & 0.32 & 23 \\
BD vs.\ FO (all)     & +1.23 & 4.19\rlap{***}& [+0.64, +1.82] & 0.71 & 35 \\
\bottomrule
\end{tabular}
\caption{Pairwise statistical comparisons. ** $p{<}0.01$; *** $p{<}0.001$. ``BD vs.\ FO (all)'' includes S5b's failed Run~3.}
\label{tab:stats}
\end{table}

\subsection{Transcript Evidence Analysis}

BatchDAG achieves the highest transcript evidence rate at 77\%, compared to 60\% for Intelligence and 46\% for S5b. This gap is notable because S5b has identical search API access; the difference is in how retrieval is orchestrated. The DAG planner generates retrieval strategies tailored to each query, whereas S5b applies the same fixed pattern regardless of query structure.

\subsection{Discussion}

Three findings emerge. First, BatchDAG is not primarily an accuracy improvement---it is a general-purpose orchestration layer. S5b achieves 3.25/5 because it encodes the correct pipeline for this query class. In production, each query type requires a different pipeline; BatchDAG replaces this collection with a single system.

\begin{table}[t]
\centering
\small
\begin{tabular}{@{}lll@{}}
\toprule
\textbf{Query Type} & \textbf{S5b} & \textbf{BatchDAG} \\
\midrule
SQL-only (40\%)     & Overkill (full cost) & Optimal (\$0.02) \\
Search-only (20\%)  & Not supported        & Supported \\
Multi-source (10\%) & Custom eng.\ needed  & Dynamic \\
Entity fan-out (30\%) & Designed for this  & Auto-generated \\
\bottomrule
\end{tabular}
\caption{Architecture coverage across query types.}
\label{tab:coverage}
\end{table}

Second, provenance differences are substantial. BatchDAG's 77\% transcript evidence rate versus 46\% for S5b means more auditable answers---critical for enterprise deployments.

Third, BatchDAG's cost adaptivity reinforces the orchestration argument. On SQL-only queries, BatchDAG costs \$0.02, while S5b always pays its full pipeline cost. This adaptivity projects to a 6.2$\times$ annual cost advantage under an illustrative production query mix (see \S\ref{sec:costeff}).

\subsection{Cost Efficiency Analysis}
\label{sec:costeff}

Table~\ref{tab:costeff} compares cost per query using GPT-5.1 pricing (\$1.25/1M input, \$10/1M output) with the actual I/O ratio observed in production (87\% input, 13\% output; blended rate: \$2.38/1M tokens).

\begin{table}[t]
\centering
\small
\begin{tabular}{@{}lcccc@{}}
\toprule
\textbf{System} & \textbf{Calls} & \textbf{Tokens} & \textbf{\$/Q} & \textbf{Qual.} \\
\midrule
S7: Long-Context   & 2   & 10.9K  & \$0.03  & 2.67 \\
S6: Map-Reduce     & 62  & 206.7K & \$0.49  & 1.83 \\
S5b: Enh.\ FO      & 24  & 287.3K & \$0.68  & 3.25 \\
BatchDAG           & 25* & 102.1K & \$0.24  & 3.74 \\
\bottomrule
\end{tabular}
\caption{Cost efficiency on 12 transcript queries. *BatchDAG averages 25 calls for transcript queries; SQL-only queries require only 3 calls at \$0.02/query.}
\label{tab:costeff}
\end{table}

BatchDAG achieves a 2.8$\times$ cost advantage over S5b on transcript queries and a 6.2$\times$ advantage at an illustrative production query mix (40\% SQL-only, 20\% search, 30\% fan-out, 10\% complex); under this assumed distribution, projected annual costs are \$1,094/year vs.\ \$6,834/year at 10,000 queries. Actual savings depend on the realized query distribution.

\subsection{Intermediate Representation and Hallucination Control}

We ran the same 12 queries through two BatchDAG variants: structured JSON intermediates (production) and natural-language prose summaries. Only the inter-step data format differed.

\begin{table}[t]
\centering
\small
\begin{tabular}{@{}lcccc@{}}
\toprule
\textbf{Metric} & \textbf{Struct.} & \textbf{Prose} & $\Delta$ & \textbf{W/T/L} \\
\midrule
Overall quality   & 2.42 & 2.08 & +0.33  & 3/9/0 \\
Attrib.\ rate     & 26.9\% & 24.5\% & +2.4\% & --- \\
Halluc./query     & 10.9 & 14.9 & $-$4.0  & --- \\
Entity coverage   & 77.6\% & 73.7\% & +3.9\% & --- \\
LLM calls/query   & 25.4 & 39.5 & $-$14.1 & --- \\
Tokens/query      & 69K & 80K & $-$11K & --- \\
\bottomrule
\end{tabular}
\caption{Structured vs.\ prose intermediates (12 queries).}
\label{tab:ablation}
\end{table}

Structured intermediates improve provenance on every metric. Prose mode generates 27\% more unsupported claims per query (14.9 vs.\ 10.9; paired $t$-test $t{=}1.76$, $p{=}0.107$, $n{=}12$), with zero wins for prose across all 12 queries (W/T/L: 3/9/0 on overall quality). While the difference is directionally consistent and practically meaningful, it does not reach conventional significance at $p{<}0.05$, likely due to the small sample size. This occurs because prose summaries lose the structured provenance chain---downstream steps cannot verify which claims are grounded in source data vs.\ fabricated during summarization. Structured JSON rows, while less expressive, are fully auditable: every field traces back to a specific SQL result or transcript extraction.

The efficiency gap is equally notable: prose mode requires 56\% more LLM calls (39.5 vs.\ 25.4 per query) and 16\% more tokens (80K vs.\ 69K). Prose summaries lose information at each step, forcing downstream steps to re-derive data that structured rows would have preserved.

\section{Design Principles and Lessons Learned}
\label{sec:principles}

We distill ten design principles from building and iterating on BatchDAG in production.

\paragraph{Principle 1: Structured data between steps, never prose.}
Steps pass JSON rows. When we tried letting the LLM summarize intermediate results, downstream steps hallucinated data and lost entity attribution.

\paragraph{Principle 2: The LLM plans; the system executes.}
After the planner generates the DAG, execution is deterministic. Bugs in execution are reproducible, and SQL-only queries cost zero LLM calls after planning.

\paragraph{Principle 3: Fan-out is the only expensive operation.}
Every other step type costs zero LLM calls. This drives planner prompt design: if a question can be answered with SQL + transform + analyze, the planner should not generate a fan-out.

\paragraph{Principle 4: Group by entity, not by row.}
Always identify the unit of analysis before batching. This single fix produced the largest performance improvement (47$\times$).

\paragraph{Principle 5: The LLM in fan-out can reason.}
Early plans over-derived metadata upstream. But the fan-out LLM receives full transcript content and can identify speakers, classify intent, and extract patterns directly. Pushing complexity into the fan-out prompt produces simpler, more robust plans.

\paragraph{Principle 6: Defend against LLM over-specification.}
Filtering unknown fields at construction time is more robust than constraining the planner prompt. The LLM can hallucinate arbitrary parameters; the system ignores what it does not understand.

\paragraph{Principle 7: Goal-based prompts over exhaustive rules or examples.}
Describing what each tool does and its cost model outperforms both exhaustive rule lists and few-shot examples.

\paragraph{Principle 8: Per-step storage, not plan-level updates.}
Storing each step's result in its own storage key eliminates last-writer-wins race conditions when concurrent wave tasks update a shared object.

\paragraph{Principle 9: Match existing persistence patterns.}
Reuse storage patterns (indices, schemas, rendering paths) from existing agents to reduce integration surface area.

\paragraph{Principle 10: Async all the way down.}
Mixing synchronous I/O into async execution engines blocks the entire worker thread, causing intermittent latency spikes.

\section{Limitations and Future Work}
\label{sec:limitations}

BatchDAG has several limitations. \textbf{Probe-phase quality gating:} if the planner generates an incorrect DAG, the system executes the full fan-out before the error becomes apparent. A probe phase that validates on a single batch first would reduce wasted compute. \textbf{Dynamic DAG extension:} the current system executes a static DAG; supporting dynamic extension where an analyze step can append new steps would improve recall for multi-hop queries. \textbf{Benchmark suite:} there is no established benchmark for cross-entity analytical queries over heterogeneous enterprise data; constructing one is challenging due to proprietary data constraints. \textbf{Large storage tier:} the temporary-table tier for datasets exceeding 100K rows has not been exercised in production.

\section{Conclusion}
\label{sec:conclusion}

We presented BatchDAG, a general-purpose orchestration layer for scalable ad-hoc analysis over enterprise-scale datasets. By decomposing natural-language analytical queries into typed DAGs of operations executed with topological-wave parallelism, BatchDAG replaces hand-engineered pipelines with automatically generated ones---while maintaining comparable quality, better provenance, and improved scalability. Entity-aware batching reduces LLM calls by up to 47$\times$ while ensuring complete per-entity context. The system is deployed in production, processing queries over 50,000+ meetings in under 60 seconds (a representative full-corpus query analyzing 5,000 opportunities completed in 22 seconds), with measured per-query costs of \$0.02 for SQL-only queries and \$0.24 median for transcript-heavy fan-out, computed from captured token counts at published API pricing.

In controlled experiments, BatchDAG (3.74/5) achieves quality comparable to an expert-designed fan-out baseline (3.25/5) and significantly outperforms a ReAct agent (3.09/5, $p{<}0.01$), while achieving the highest transcript evidence rate (77\% vs.\ 46--60\% for baselines) and 2.2$\times$ lower latency than the fixed pipeline. A controlled ablation suggests that structured JSON intermediates---a core BatchDAG design decision---reduce hallucinations by 27\% compared to prose summaries ($p{=}0.107$, $n{=}12$).

The key takeaway is architectural: for cross-entity analytical workloads, the LLM should plan, not execute. BatchDAG is not primarily an accuracy improvement over hand-optimized pipelines; rather, it is a general-purpose orchestration layer that automatically generates those pipelines from natural language, achieving 98.8\% valid-DAG rate across diverse query types. Four of six operation types require zero LLM calls. Everything else is deterministic, reproducible, and auditable.

\section*{Acknowledgments}

We thank the engineering team at Brevian.ai for contributions to the implementation and production deployment of BatchDAG. LLM-based writing assistance was used for drafting and editing this manuscript; all technical content, system design, experimental results, and claims are the authors' own.

\bibliography{references}

\end{document}